\documentclass[conference]{IEEEtran}
\IEEEoverridecommandlockouts
\usepackage{cite}
\usepackage{amsmath,amssymb,amsfonts}
\usepackage{algorithmic}
\usepackage{graphicx}
\usepackage{textcomp}
\usepackage{xcolor}
\usepackage{amsmath,lipsum}
\usepackage[ruled]{algorithm2e}
\usepackage{graphicx}
\usepackage{booktabs}
\usepackage{float}
\usepackage{subfigure}
\usepackage{multirow}
\usepackage{multicol}
\usepackage{arydshln}
\usepackage{cite}
\usepackage{stfloats}
\def\BibTeX{{\rm B\kern-.05em{\sc i\kern-.025em b}\kern-.08em
    T\kern-.1667em\lower.7ex\hbox{E}\kern-.125emX}}
\begin{document}

\title{Semantic Communication Enabling Robust Edge Intelligence for Time-Critical IoT Applications\\ }
\author{\IEEEauthorblockN{Andrea~Cavagna, Nan~Li, Alexandros~Iosifidis and Qi~Zhang }
\IEEEauthorblockA{DIGIT, Department of Electrical and Computer Engineering, Aarhus University.\\
Email: andrea.cavagna.94@gmail.com, \{linan, ai, qz\}@ece.au.dk}}
\maketitle

\begin{abstract}
This paper aims to design robust Edge Intelligence using semantic communication for time-critical IoT applications. We systematically analyze the effect of image DCT coefficients on inference accuracy and propose the channel-agnostic effectiveness encoding for offloading by transmitting the most meaningful task data first. This scheme can well utilize all available communication resource and strike a balance between transmission latency and inference accuracy. Then, we design an effectiveness decoding by implementing a novel image augmentation process for convolutional neural network (CNN) training, through which an original CNN model is transformed into a Robust CNN model. We use the proposed training method to generate Robust MobileNet-v2 and Robust ResNet-50. The proposed Edge Intelligence framework consists of the proposed effectiveness encoding and effectiveness decoding. The experimental results show that the effectiveness decoding using the Robust CNN models perform consistently better under various image distortions caused by channel errors or limited communication resource. The proposed Edge Intelligence framework using semantic communication significantly outperforms the conventional approach under latency and data rate constraints, in particular, under ultra stringent deadlines and low data rate.
\end{abstract}


\section{Introduction}

Artificial Intelligence based on convolutional neural networks (CNN) is widely applied for Internet of Things (IoT) applications \cite{Marculescu2020}. For time-critical IoT applications, e.g. autonomous driving, industrial automation and control, and smart surveillance,  it is indispensable to provide services not only of high reliability (e.g. high inference accuracy) but also within ultra-low
latency~\cite{Liu2020}. However, it is challenging for resource-constrained
IoT devices to run computation-intensive inference tasks in real time locally \cite{Nan2022}.

Edge Computing is a promising approach to address this challenge, in which IoT
devices offload intensive computation tasks to edge servers
(ESs) in their proximity \cite{Guo2022}. Most of the existing
works in Edge Computing focus on reliable transmission in offloading (i.e. accurately receiving the transmitted bits/packets) instead of service reliability \cite{Liu2020IOT}.
The work \cite{Nan2022} show that it is more important to ensure service
reliability from an application's perspective, i.e. to provide good inference accuracy within the deadline. This is aligned with the emerging paradigm shift from “semantic neutrality” in the conventional communication based on Shannon's Information Theory towards \textit{semantic communication}~\cite{QiaoSemetic, Weng2021}.

Semantic communication aims to achieve that the transmitted data can precisely convey the desired meaning and the received data enable the receiver to achieve a final goal using the received data, e.g. to perform inference or make the right decision, instead of simply focusing on reliable data transmission. A semantic communication system is usually composed of semantic/effectiveness encoding and semantic/effectiveness decoding. Effectiveness encoding is the process to encode large raw data (e.g. high-dimensional data) into more compact data representation (e.g. reduced-dimension features)~\cite{Luo2022}. In the context of Edge Intelligence, effectiveness decoding basically includes receiving task data, decoding the data (e.g. image reconstruction), and performing inference at Edge server~\cite{Shao2022}.

In time-critical IoT applications, on the one hand, it needs to decide how much data can be transmitted within limited communication time budget, furthermore, retransmission is often not feasible in offloading due to the excessive transmission latency it causes. On the ohter hand, as we know that slight distortion in input images will not cause significant loss of inference accuracy~\cite{Xiao_2022_CVPR, Wang2022IoT}. For example, the face recognition can still achieve acceptable inference accuracy as long as less than 13.3\% of the image pixels are corrupted~\cite{Wright2009Robust}. This shows that inference accuracy is not fully dependent on bit-level transmission reliability. Therefore, it is worth studying how to design an effectiveness encoding that uses less communication resource (e.g. transmission time, bandwidth) and how to design the corresponding effectiveness decoding that consistently achieves high inference accuracy.

In this paper, we aim to design an Edge Intelligence framework using semantic communication paradigm, which can enhance CNN-based inference at ES within strict time constraints, while ensuring decent inference accuracy even under severe channel errors without retransmission. In particular, the key research questions are in the following.
\begin{itemize}
  \item In general, high-quality input image generates high inference accuracy, whereas it leads to longer transmission latency. How can we design an effectiveness encoding to reduce transmission data size and shorten transmission latency while ensuring inference accuracy?
  \item	Stochastic wireless communication channel can cause bit errors and even packet losses \cite{Hentati2020}. Standard error control schemes, Automatic Repeat Request (ARQ), can incur excessive transmission latency. Forward Error Control (FEC) of strong error correction capability can add long parity check bits and consequently increase transmission latency. How can we design an effectiveness decoding to ensure inference accuracy without using ARQ-based retransmission nor FEC with low code rate?
\end{itemize}

To address these questions, we encode an image using its discrete cosine transform (DCT) representation~\cite{Shen_2021_CVPR} and the DCT coefficients are transmitted in sequence based on their importance. The effectiveness decoder can reconstruct an image from any subset of its original DCT representation. The decoder is designed in such a way that it can dynamically terminate reception if the communication time budget is over and start inference using all received DCT coefficients. To ensure inference accuracy without retransmission or with low-quality image, we design the effectiveness decoder by designing a novel CNN training method to cope with images of few DCT coefficients and images containing bit errors or packet losses. The main contributions are summarized as follows.
\begin{itemize}
   \item We systematically analyze the effect of missing DCT coefficients on image classification, and propose the channel-agnostic effectiveness encoding for offloading, which can make a flexible utilization of all available communication resource and strike a good balance between transmission latency and inference accuracy.
  \item	We implement the effectiveness decoding through a novel image augmentation process for CNN training. It transformed original CNN models into new Robust CNN models that can provide high inference accuracy without all DCT coefficients. We use this training method to generate robust versions of MobileNet-v2~\cite{Sandler2018MobileNetV2} and ResNet-50~\cite{He2016resnet50}. The experimental results show that the Robust CNN models outperform their original models under missing DCT coefficients. For example, effectiveness decoding Robust ResNet-50 achieves a gain of inference accuracy up to 17.9\% on flower dataset \cite{Alexander2021} and 25.3\% on Caltech 256 \cite{caltech256}.
  \item	To demonstrate the feasibility of our designed effectiveness decoding using the Robust CNN models, we test their performance in an error-prone communication channel with different bit error probabilities and packet loss rates. The experimental results show that the Robust CNN models can perform better than their original models, in particular, when packet loss rate is high. For example, our Robust MobileNet-V2 achieves a gain of average inference accuracy up to 11.2\% on flower dataset and 20.4\% on Caltech 256 when 5 packets in each image are lost. For the case of 10\% uniformly distributed bit errors, Robust ResNet-50 achieves a gain of average inference accuracy up to 24.3\% for flower dataset and 38.3\% for Caltech 256. We also apply typical effectiveness encoding strategies to combat bit errors and it is possible to achieve extra 16.1\% inference accuracy improvement for flower dataset using Robust ResNet-50. 
  \item We further study the performance of our proposed Edge Intelligence framework using semantic communication under the latency and transmission rate constraints. The experimental results show that it outperforms the conventional framework, achieving a gain up to 40\% for flower dataset using Robust ResNet-50. In particular, under extremely stringent deadlines and low data rates, the proposed framework can still achieve decent inference accuracy whereas the conventional approach will completely fail.
\end{itemize}

The remainder of this paper is organized as follows. Section \ref{Section:CNNTraining} proposes channel-agnostic effectiveness encoding for offloading to strike a balance between transmission latency and inference accuracy, and describes the training method of Robust CNN which can deal with missing DCT coefficients. The experimental results are presented and discussed in Section~\ref{Section:PerformanceEva}, and the conclusions are drawn in Section \ref{Section:Conclusion}.

\begin{figure*}
\centering
\includegraphics[width=0.9\textwidth]{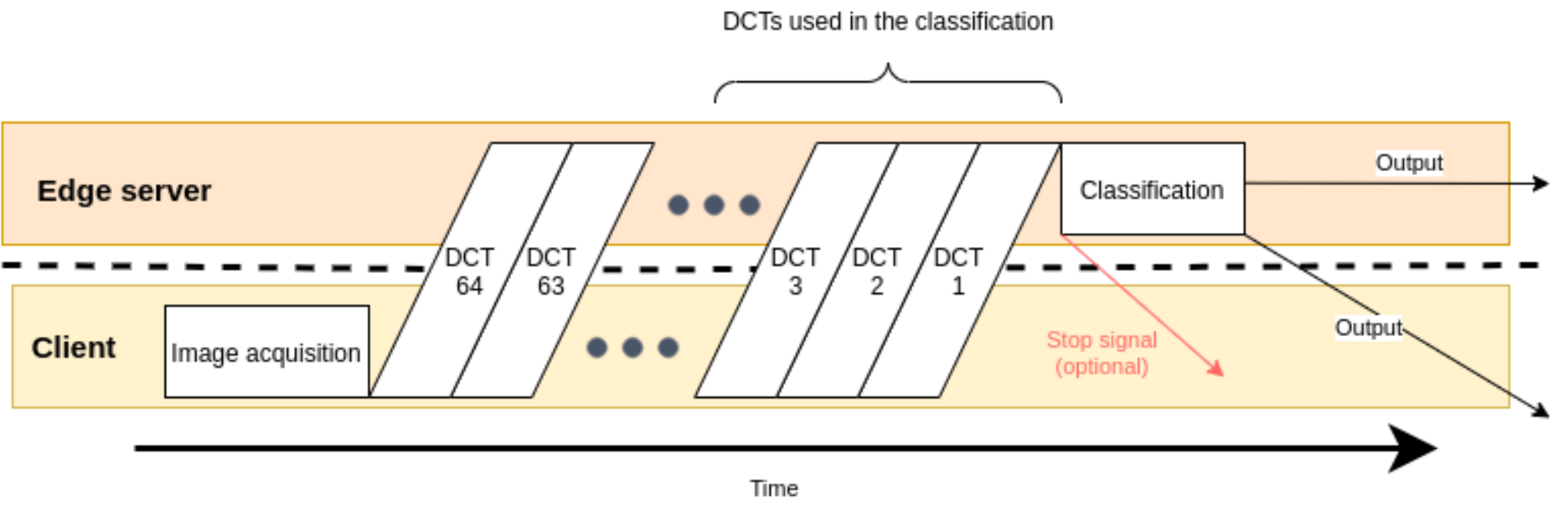}
\caption{Channel-agnostic effectiveness encoding for offloading.}
\label{fig:coeff}
\end{figure*}

\section{Edge Intelligence Using Semantic Communication}\label{Section:CNNTraining}
This section describes the design and development of the proposed Edge Intelligence with semantic communication. To ensure the inference accuracy within a latency constraint, it is important to make use of the communication time budget to receive the best possible representation of an image, regardless of wireless channel state. Therefore, we design an effectiveness encoding which can perform fast image encoding for efficient computation offloading, and an effectiveness decoding which can dynamically terminate the reception and start image reconstruction and inference. Furthermore, we design and implement the effectiveness decoding in such a way that it still achieves high inference accuracy under missing DCT coefficients. 
\subsection{Channel-agnostic Effectiveness Encoding for Offloading}\label{Subsec:OffloadingScheme}
In conventional offloading scheme for applications with hard deadline, an IoT device first estimates the available time budget for data transmission based on the average task computation time and it calculates the amount of data that can be transmitted within the time budget based on channel estimation. Then, an IoT device can select proper compression parameters, for example using JPEG, to ensure completing inference at ES within the latency constraint. However, due to time-varying wireless channel, continuous channel estimation is required to do this. In addition, the actual data rate and channel bit error probability might not be accurately estimated. Furthermore, JPEG has limited compression options, and in some cases the minimal size of a compressed image may still be larger than the allowed data size, which can cause significant distortions in a JPEG image.

To address these issues, we design a channel-agnostic effectiveness encoding for offloading, as shown in Fig.~\ref{fig:coeff}. The basic idea is to let IoT device transmit as many DCT coefficients as possible without the knowledge of communication channel state and allow ES to decide when to start inference. This design is based on the fact that each DCT coefficient makes an additive contribution to the received image quality. 

The rationale behind this encoding design is as follows. Each DCT coefficient encodes frequency information of an $8\times8$ block in an image. This means that every received DCT coefficient contributes information to the entire block. Using this type of compression will significantly reduce the size of a compressed image, especially in blocks with fairly uniform luminance or chrominance (i.e. most of the DCT coefficients are close to zero). Thus, the received image gradually gets refined after receiving more and more DCT coefficients within the allowed communication time. As IoT device and ES are synchronized, an ES can start to decode the received DCT coefficients and classify the reconstructed image as soon as the communication time budget is over. In this way, it is able to provide a representation of the input image by fully utilizing the available communication resource. Furthermore, it avoids the processing, such as channel estimation and selecting compression parameters, thereby reducing energy consumption at IoT device.
\begin{figure}
\centering
\centering
       \subfigure[Remove 1st]{
		\includegraphics[width=0.11\textwidth]{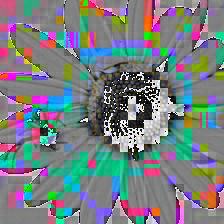}}
		\subfigure[Remove 2nd]{
		\includegraphics[width=0.11\textwidth]{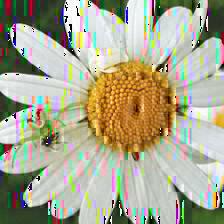}}
		\subfigure[Remove 3rd]{
		\includegraphics[width=0.11\textwidth]{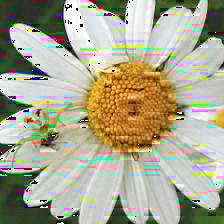}}
		\subfigure[Remove 5th]{
		\includegraphics[width=0.11\textwidth]{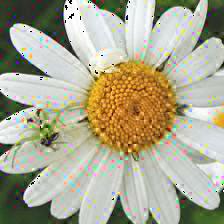}}
\caption{Remove the $n$-th DCT coefficients}
\label{fig:remove}
\end{figure}

\begin{figure}
\centering
	\subfigure[No loss in first two packets]{{\includegraphics[width=0.25\linewidth]{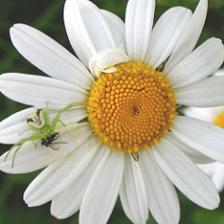} }}%
	\hspace{0.4cm}
	\subfigure[1st packet lost ]{{\includegraphics[width=0.25\linewidth]{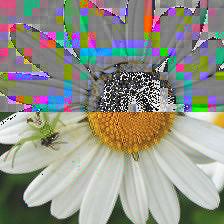} }}%
	\hspace{0.4cm}
	\subfigure[2nd packet lost ]{{\includegraphics[width=0.25\linewidth]{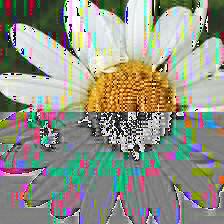}} }%
	\caption{Images with a single packet loss}%
	\label{fig:packetlost2}%
\end{figure}
\begin{figure}[t]
\centering
       \subfigure[Keep top-1]{
		\includegraphics[width=0.11\textwidth]{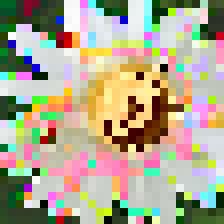}}
		\subfigure[Keep top-5]{
		\includegraphics[width=0.11\textwidth]{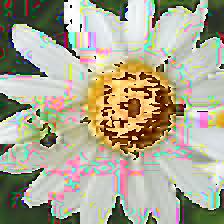}}
		\subfigure[Keep top-10]{
		\includegraphics[width=0.11\textwidth]{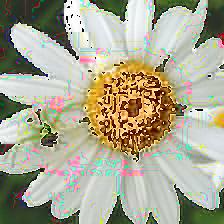}}
		\subfigure[Keep top-20]{
		\includegraphics[width=0.11\textwidth]{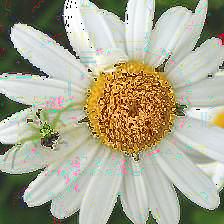}}
\caption{Keep the top-$n$ DCT coefficients}
\label{fig:keep}
\end{figure}

To better illustrate the importance of different DCT coefficients for image quality, we visually present the reconstructed images under different scenarios of missing DCT coefficients. The DCT coefficients' importance is ordered based on JPEG ZigZag matrix \cite{jpeg2015}. To achieve the best possible inference accuracy with limited communication time for offloading, it is better to transmit the most important DCT coefficients first. As shown in Fig.~\ref{fig:remove}, the loss of the 1st DCT coefficient results in severe image distortion. Note that for the flower dataset all DCT coefficients of an image are packetized into 36 packets and the first 2 packets contain all the 1st DCT coefficients. Fig.~\ref{fig:packetlost2} shows the distorted images due to one packet loss. We can see that it has a great impact on a reconstructed image, if a packet containing the 1st DCT coefficients is lost. Fig.~\ref{fig:keep} shows the examples of keeping the top-$n$ coefficients of the 64 DCT coefficients. We can see that the image can be visually well recognized with merely the top-5 DCT coefficients. This shows that the loss of high frequency components does not significantly reduce inference accuracy, which is verified in Section~\ref{Section:KeepingTopN}.
\begin{figure*}
    \centering
    \includegraphics[width=0.95\textwidth]{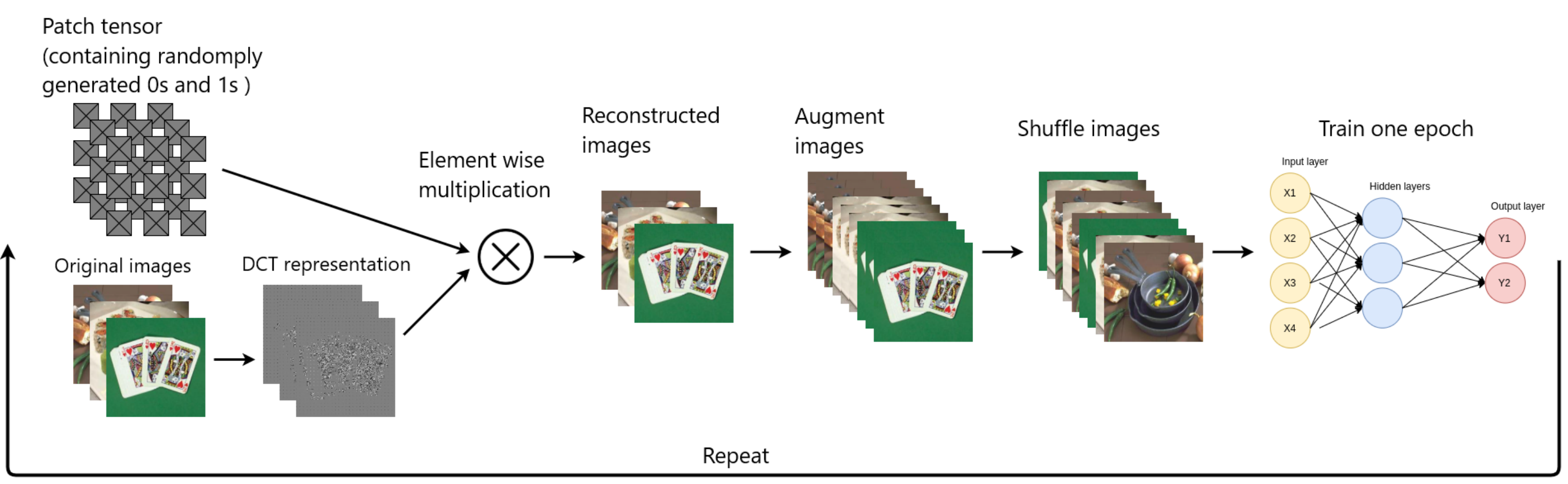}
    \caption{Robust CNN training method for varying number of DCT coefficients}
    \label{fig:training}
\end{figure*}
\subsection{Robust CNN Training Method for Effectiveness Decoding}
In the context of Edge Intelligence, the effectiveness decoding process in semantic communication is to reconstruct image from the received DCT coefficients and then perform inference, e.g. image classification, using a CNN model. The performance metric of effectiveness decoding is inference accuracy. Images containing artifacts such as pixelation and color distortion, may cause inference accuracy loss. We have observed through experiments that a conventional CNN model trained on the original images often performs poorly, i.e. inference accuracy deteriorates, when an input image has significant distortion. To meet stringent deadline of time-critical IoT applications, IoT device often needs to compress image size, furthermore, it is not viable to rely on retransmission or advanced FEC in edge computation offloading. Therefore, we aim at training a Robust CNN model for effectiveness decoding that can achieve high accuracy, regardless of the received image quality. 

We also realized that the reason of inference accuracy loss under image distortion lies in that an conventional CNN model always expects an input image of the original training image quality, consequently, the filters and parameters learned in the training struggle to recognize the same pattern in an input image even though the image has different degrees of distortions. To address this issue, during the training phase a Robust CNN model should learn more filters for the same class and even for the same image, having different degrees of distortions. In this way, the new Robust CNN model will learn new filters associated with the same class increasing the probability of a correct classification. This is possible because neural network, and especially CNN, consists of millions of parameters and so it can learn multiple representations of the same pattern within a single CNN model. Based on these hypotheses, we implement a new training technique that involves the use of imperfect images, as shown in Fig.~\ref{fig:training}. To save the time needed to generate each image with high diversity in the data set, we combine our new training technique with some standard image augmentation techniques. At each epoch of the CNN parameters optimization process, we augment the training and validation sets as follows. (i) Each image is transformed to its DCT representation. (ii) Each DCT coefficient can get lost with a given probability. (iii) Images are reconstructed by using the corrupted DCT coefficients.

The image augmentation process generates a large number of reconstructed images. The generated images are further augmented using standard geometric transformation (rotation and scaling) and used to update the CNN model parameters. In each epoch of the training process, the data enhancement process is repeated to prevent the model from overfitting image distortion in the data generated by each epoch, and to improve the inference accuracy when severely distorted images appear.

\begin{figure*}
	\centering
	\subfigure[Flower]{
		\includegraphics[width=0.45\textwidth]{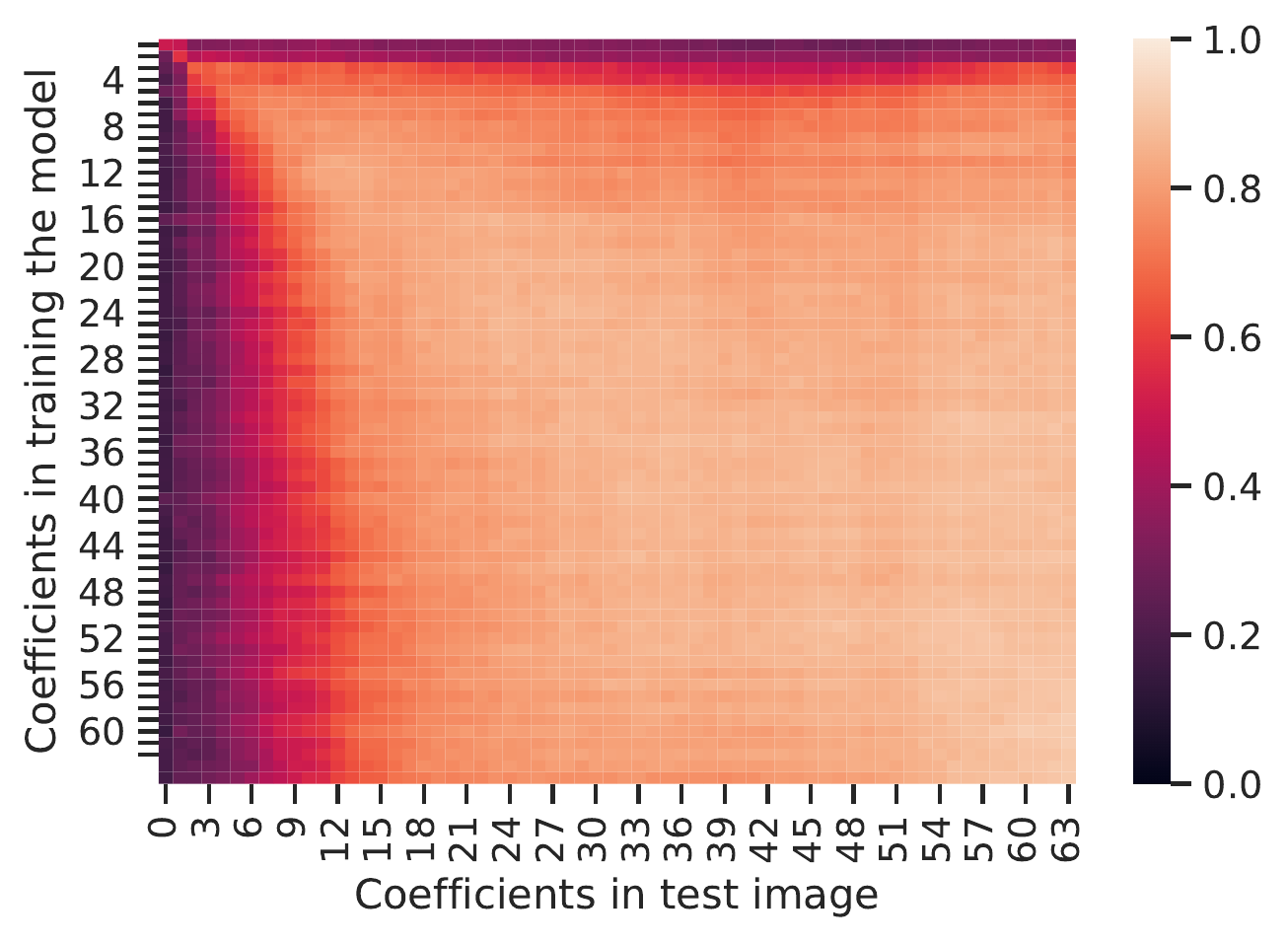}}
	\subfigure[Caltech 256]{
		\includegraphics[width=0.4\textwidth]{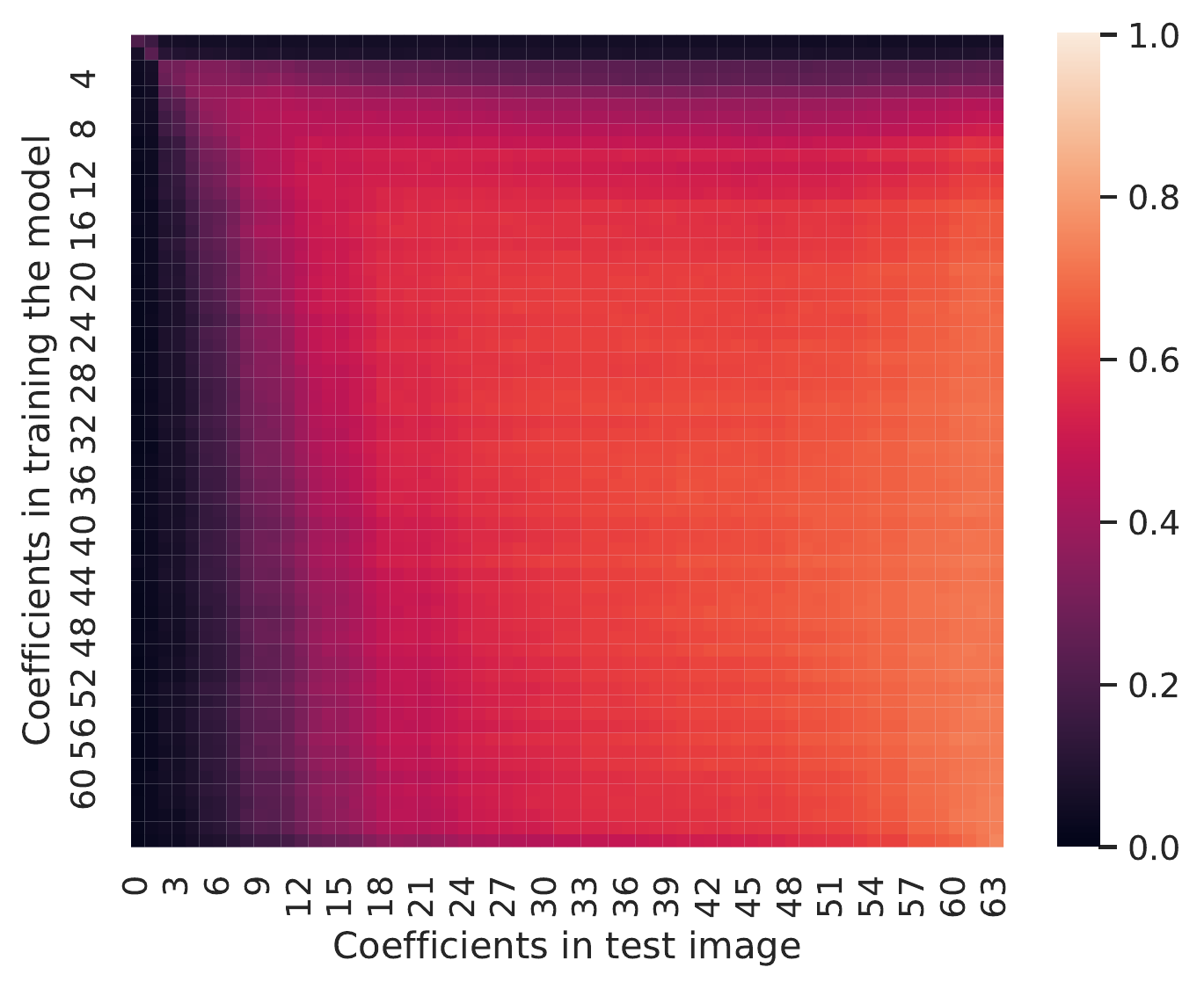}}
	\caption{Inference Accuracy heatmap of MobileNet-V2 obtained by keeping the top-$n$ coefficients on training and test images}%
	\label{fig:heatmapkeepn}%
\end{figure*}
\section{Performance Evaluation}\label{Section:PerformanceEva}
In this section we systematically studied the impact of distortion caused by missing DCT coefficients. Then we compare the inference accuracy of the proposed Edge Intelligence using semantic communication and the conventional approach (i.e., the conventional offloading using the original CNN model at Edge) under different image distortions, as well as under latency and data rate constraints. The flowers dataset~\cite{Alexander2021} composed of over 4000 images in 5 categories and Caltech 256 \cite{caltech256} composed of over 30000 images in 256 categories are used in the experiments.
\begin{table}
  \caption{Average inference accuracy by keeping top-$n$ DCT coefficients}
\label{tab:keepingTopN}
\scalebox{0.9}{
  \begin{tabular}{ccccccc}
    \toprule
    \multicolumn{1}{c}{} & \multicolumn{3}{c}{\textbf{Flower}} & \multicolumn{3}{c}{\textbf{Caltech 256}}\\
 & Original & Robust &Diff. & Original & Robust &Diff.\\ 
    \midrule
    MobileNet-v2 & 0.705 & \textbf{0.875} & + 0.170 & 0.498&
    \textbf{0.724} & + 0.226\\ 
    ResNet-50 & 0.717 &  \textbf{0.896} & + 0.179 & 0.503 &  \textbf{0.756} & + 0.253\\
  \bottomrule
\end{tabular}}
\end{table}
\subsection{Inference accuracy by keeping Top-$n$ DCT coefficients}\label{Section:KeepingTopN}
In this experiment, we studied the relationship between the model trained using distorted images and the test data set consisting of distorted images. We use the proposed training method to train MobileNet-V2 and ResNet-50 on images with different levels of image distortion, by keeping the top-$n$ DCT coefficients, where $n$ varies from 1 (heavy distortion) to 64 (no distortion). In this way, we obtained 64 models. We tested all the 64 models on the test data set with top-$n$ DCT coefficients. Fig.~\ref{fig:heatmapkeepn} shows the inference accuracy of the trained models on images with different amount of top-$n$ DCT coefficients. It shows a clear correlation between the distortion level of the training images and that of the test images, in other words, for a given distorted image higher inference accuracy is achieved, if the model was trained using distorted images with similar amount of top-$n$ DCT coefficients. 

Table~\ref{tab:keepingTopN} summarizes the average inference accuracy of Robust MobileNet-V2 and Robust ResNet-50, and their original models on distorted images (i.e. with top-$n$ DCT coefficients, $n$ varying from 1 to 64). We can see that under image distortion the effectiveness decoding using Robust MobileNet-V2 and Robust ResNet-50 achieve inference accuracy improvement of 17\% and 17.9\% on the flower dataset and 22.6\% and 25.3\% on the caltech 256, respectively. In general, the performances of Robust CNN models are much better when the images are heavily distorted (e.g.~$n < 20$), and the inference accuracy difference becomes less significant when the images have little distortion. This study illustrates the rationale behind the design of channel-agnostic effectiveness encoding for offloading. It means under stringent latency and data rate constraints, even if an IoT device is only able to transmit the top-$n$ DCT coefficients ($n$ can be very small, causing severe image distortion), the Robust CNN models are still able to achieve decent inference accuracy. The detailed results will be presented in Subsection~\ref{SubSec:delaytransmissionrateconstraints}.
\begin{table}
  \caption{Average inference accuracy under varying \textit{Q}}
\label{tab:compress_factor} 
\scalebox{0.85}{
  \begin{tabular}{ccccccc}
    \toprule
    \multicolumn{1}{c}{} & \multicolumn{3}{c}{\textbf{Flower}} & \multicolumn{3}{c}{\textbf{Caltech 256}}\\
 & Varying Q & Robust &Diff. & Varying Q & Robust &Diff.\\ 
    \midrule
    MobileNet-v2 & 0.708 & \textbf{0.741} & + 0.033 & 0.476 & \textbf{0.510} & + 0.035\\ 
    ResNet-50 & 0.777 & \textbf{0.820} & +0.043 & 0.514 & \textbf{0.577} & +0.062\\
  \bottomrule
\end{tabular}}
\end{table}
\begin{figure*}
	\centering
	\subfigure[Flower]{
		\includegraphics[width=0.4\textwidth]{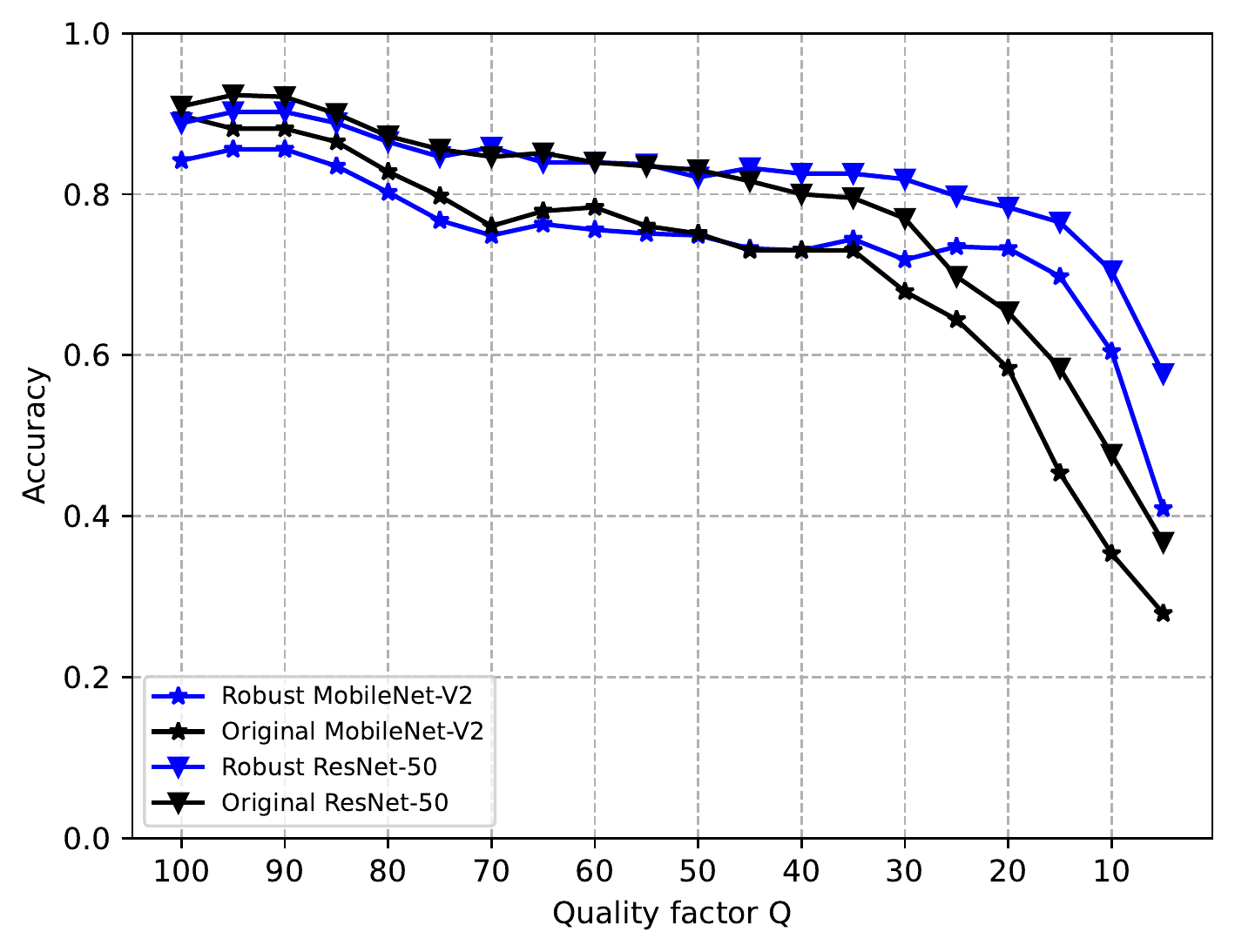}}
	\subfigure[Caltech 256]{
		\includegraphics[width=0.4\textwidth]{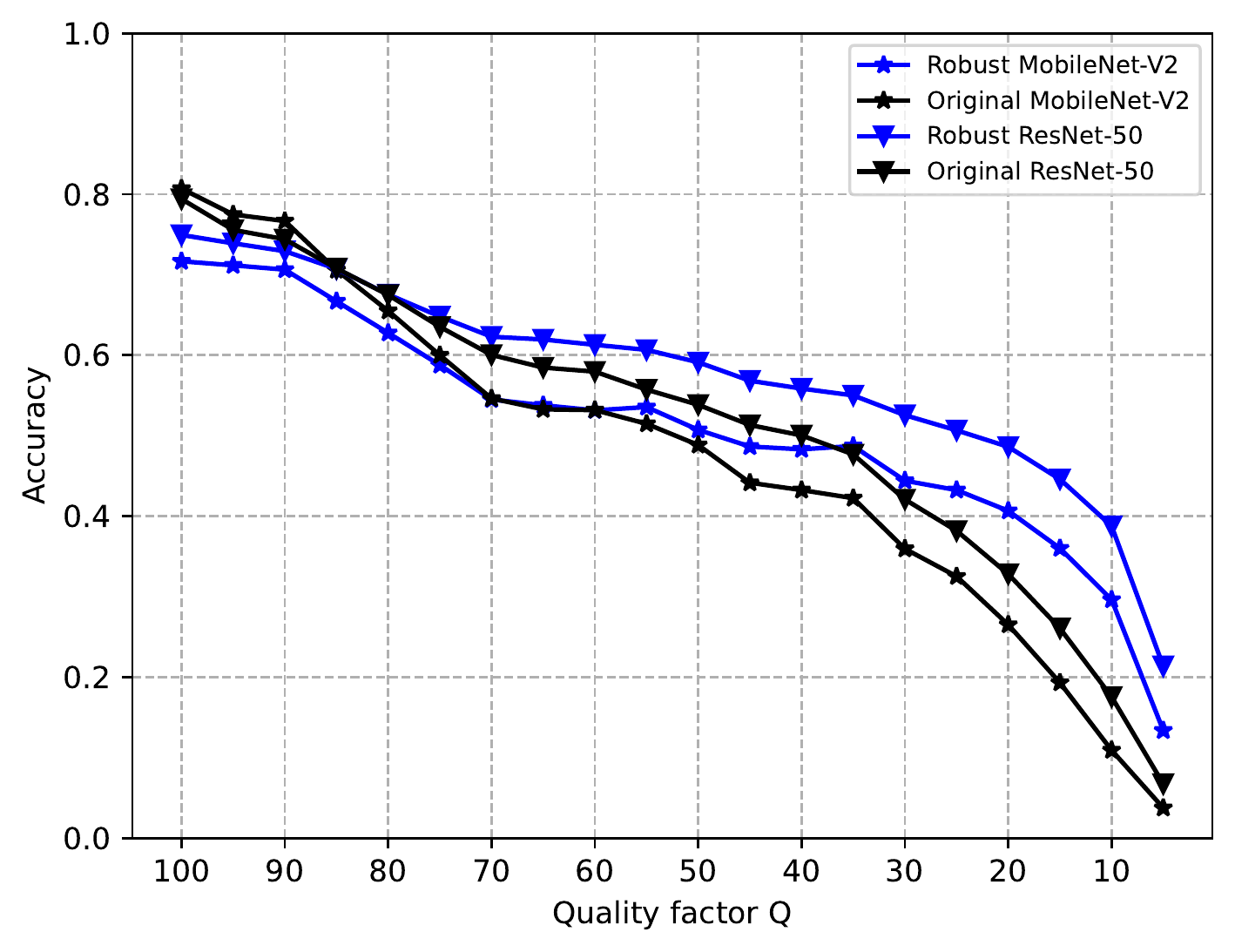}}
	\caption{Inference accuracy of Robust CNN models and their original models on distorted images under varying quality factor}%
	\label{fig:lossy_compression}%
\end{figure*}
\subsection{Inference accuracy under lossy compression}
To reduce the communication latency in computational offloading, reducing the image size by compression is a straightforward approach. In this way, it is possible to trade off inference accuracy and latency. The standard JPEG uses quality factor \textit{Q} to compress images by applying \textit{Quantization Matrix} to the the matrix of DCT coefficients. The quantization matrix follows the same principle that applying higher quantization for the less important DCT coefficient. Therefore, after quantization some of the DCT coefficients at high frequency will become zeros. The compressed image has similar data representation as our proposed compression method in Subsection~\ref{Subsec:OffloadingScheme}, the difference is that for JPEG the Q factor needs to be selected before encoding starts whereas our offloading scheme allows encoding in the air. 

Fig. \ref{fig:lossy_compression} shows the inference accuracy of our proposed Robust MobileNet-V2 and Robust ResNet-50, and their original models on distorted images under varying \textit{Q}. It can be seen that although our proposed Robust CNN models are trained on different types of images, they seem to perform better than the original models especially when the images are highly compressed. For example, when $\textit{Q}=10$, Robust ResNet 50 improve accuracy of 22.8\% (i.e., $0.705-0.477$) for flower dataset and 21.3\% (i.e., $0.388-0.175$) for Caltech 256.
This shows that our proposed Robust CNN models not only performs better for images with few DCT coefficients but also for highly compressed images with severe quantization errors. Table~\ref{tab:compress_factor} summarizes the average inference accuracy of Robust MobileNet-V2 and Robust ResNet-50, and their original models on distorted images under varying \textit{Q}. We can see that Robust MobileNet-V2 and Robust ResNet-50 outperform their original models for both flower dataset and Caltech 256. For example, Robust ResNet50 improves average accuracy by 4.3\% and 6.2\% for Flower dataset and 
Caltech 256, respectively. 
\begin{figure*}
	\centering
	\subfigure[Flower]{
		\includegraphics[width=0.4\textwidth]{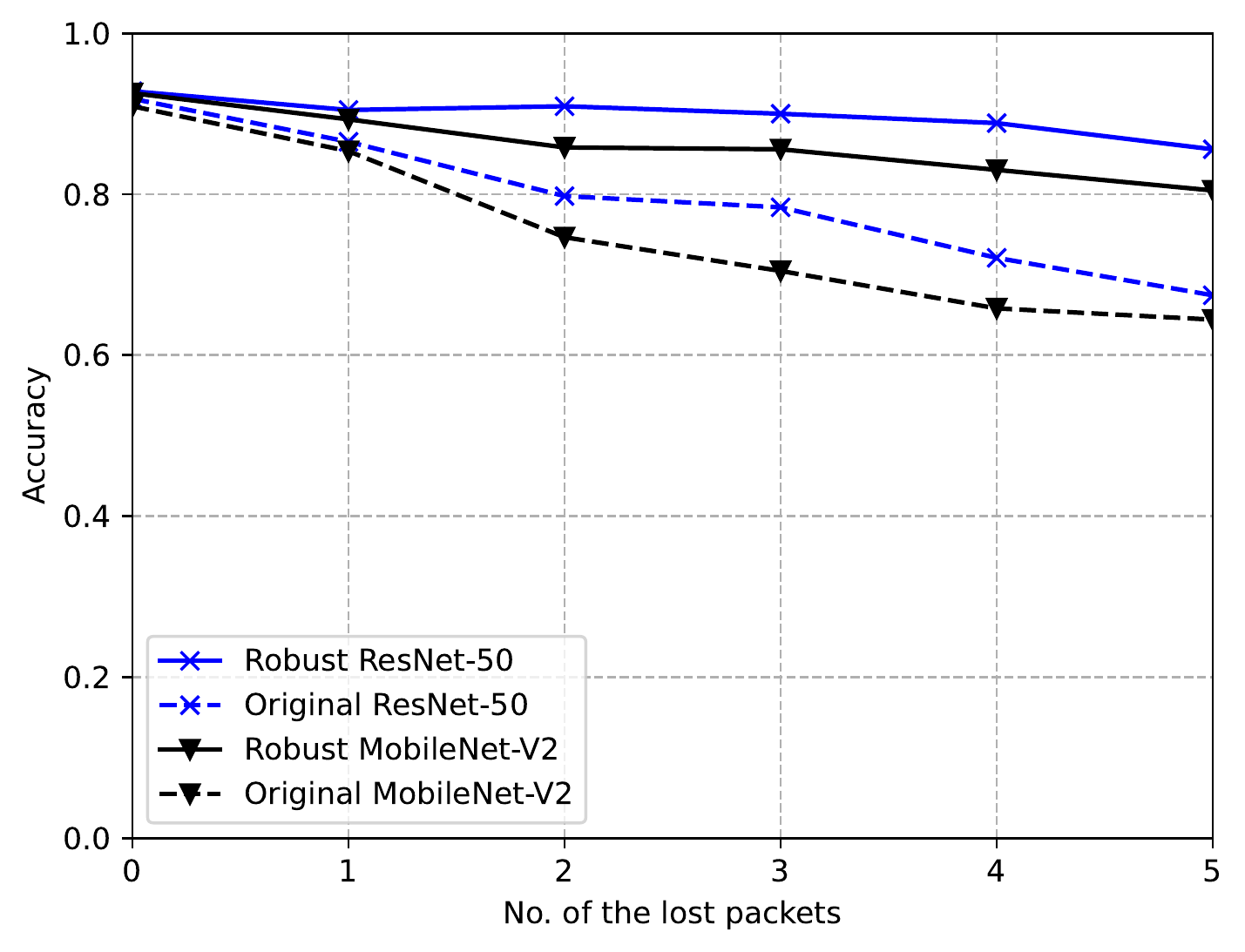}}
	\subfigure[Caltech 256]{
		\includegraphics[width=0.4\textwidth]{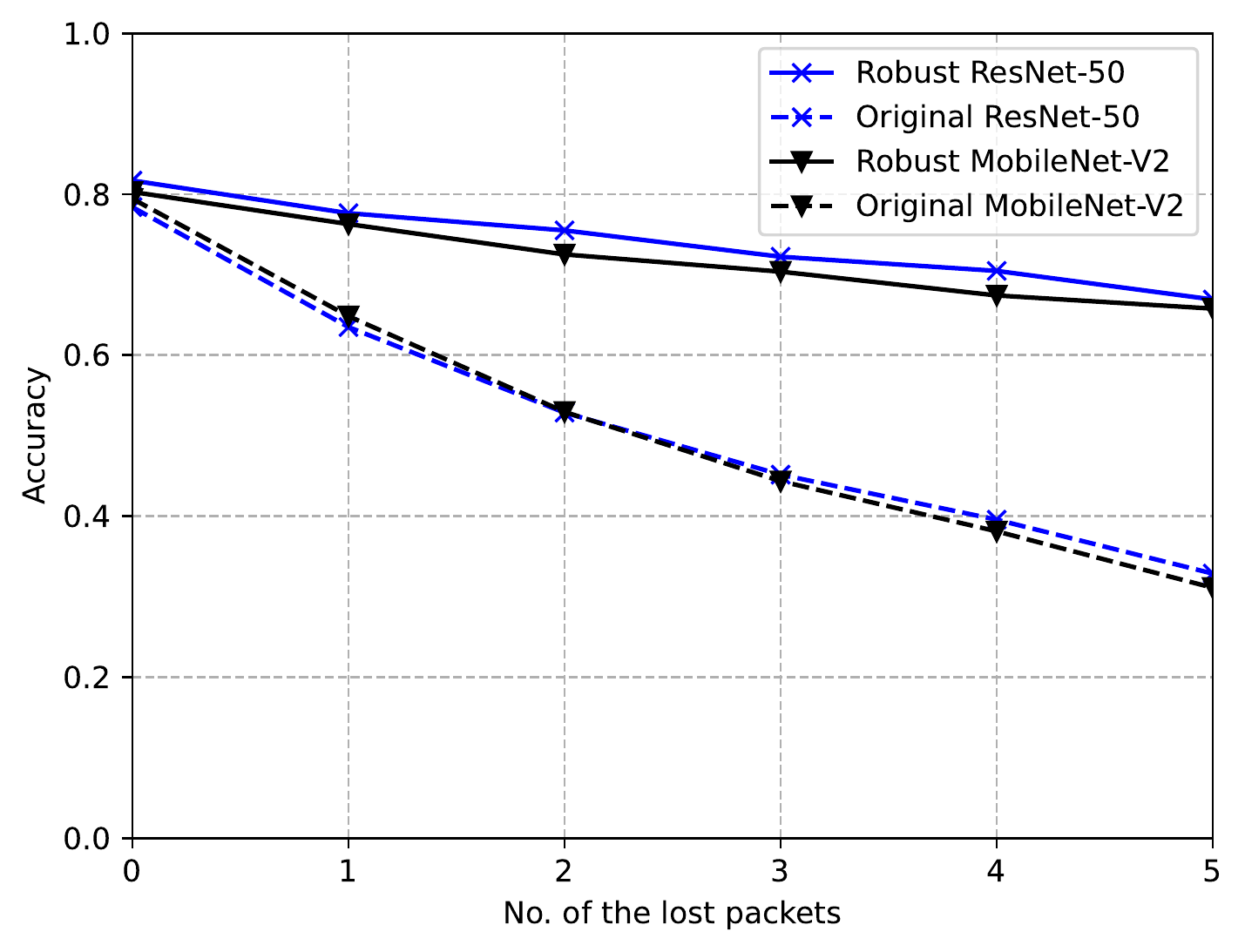}}
    \caption{Inference accuracy under different packet losses}
	\label{fig:packetdrop}
\end{figure*}
\subsection{Inference accuracy under packet losses}
For real-life IoT applications, DCT coefficients are packetized and therefore loss of DCT coefficients is, de factor, correlated to packet loss. In this experiment, we study the impact of packet losses. One packet loss may cause one or more DCT coefficients losses, depending on the packet size and the amount of data of a DCT coefficient. 
Note that the higher order DCT coefficients have smaller values and are represented by fewer bits. Assuming that packet size is of 1024 bytes, which is approximately half
of the data size of the 1st DCT coefficients in all blocks, namely, the 1st DCT coefficients of all blocks in an image can be transmitted in two packets. In the experiment, we measured the inference accuracy when each image randomly loses 1 to 5 packets.
\setcounter{table}{2}
\begin{table}
  \caption{Average inference accuracy under 0-5 packet losses}
\label{tab:Packetloss} 
\scalebox{0.9}{
  \begin{tabular}{ccccccc}
    \toprule
    \multicolumn{1}{c}{} & \multicolumn{3}{c}{\textbf{Flower}} & \multicolumn{3}{c}{\textbf{Caltech 256}}\\
 & Original & Robust &Diff. & Original & Robust &Diff.\\ 
    \midrule
    MobileNet-v2 & 0.751 & \textbf{0.863} & + 0.112 & 0.517 & \textbf{0.721} & + 0.204\\ 
    ResNet-50 & 0.793 & \textbf{0.897} & + 0.104 & 0.520 &  \textbf{0.740} & + 0.220\\
  \bottomrule
\end{tabular}}
\end{table}

Fig.~\ref{fig:packetdrop} compares the inference accuracy of Robust CNN models and their original models, when 0 to 5 packets are lost. It can be seen that the effectiveness decoding using Robust CNN models achieves 
higher inference accuracy than the original models.
Table~\ref{tab:Packetloss} summarizes the average inference accuracy of the Robust CNN models and their original models for MobileNet-V2 and ResNet-50 under packet losses from 0 to 5. We can see a similar performance improvement by using Robust CNN models. For example, Robust ResNet-50 achieves an average inference accuracy of 89.7\% (i.e. $\left(0.927+0.904+0.909+0.899+0.888+0.855\right)/6$) on flower dataset  and 74.0\% on Caltech 256
(i.e. $
\left(0.817+0.776+0.755+0.722+0.704+0.669\right)/6$), which are 10.4\% and 22.0\% higher than that of their original models. 
\begin{figure*}
	\centering
	\subfigure[Flower]{
		\includegraphics[width=0.4\textwidth]{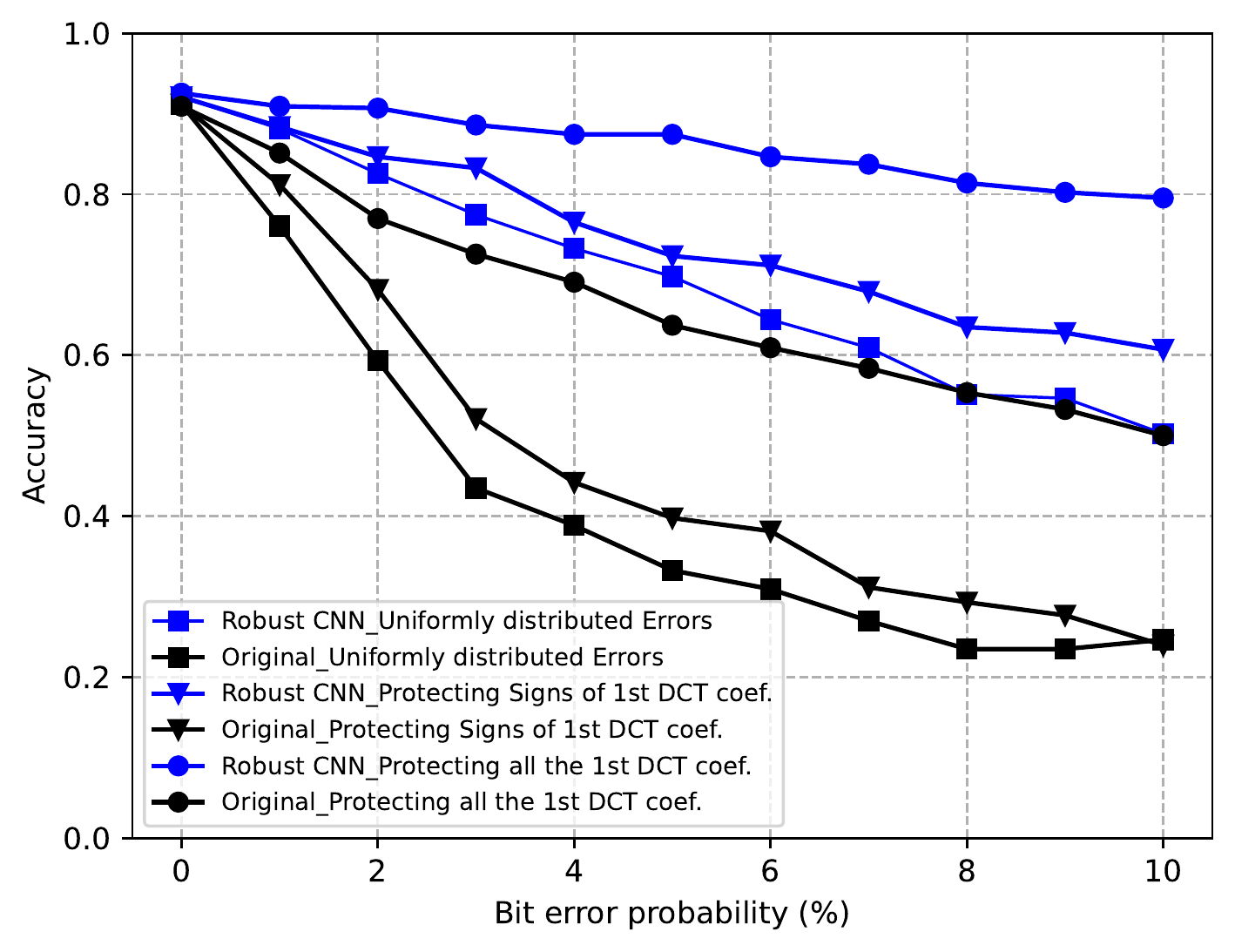}}
	\subfigure[Caltech 256]{
		\includegraphics[width=0.4\textwidth]{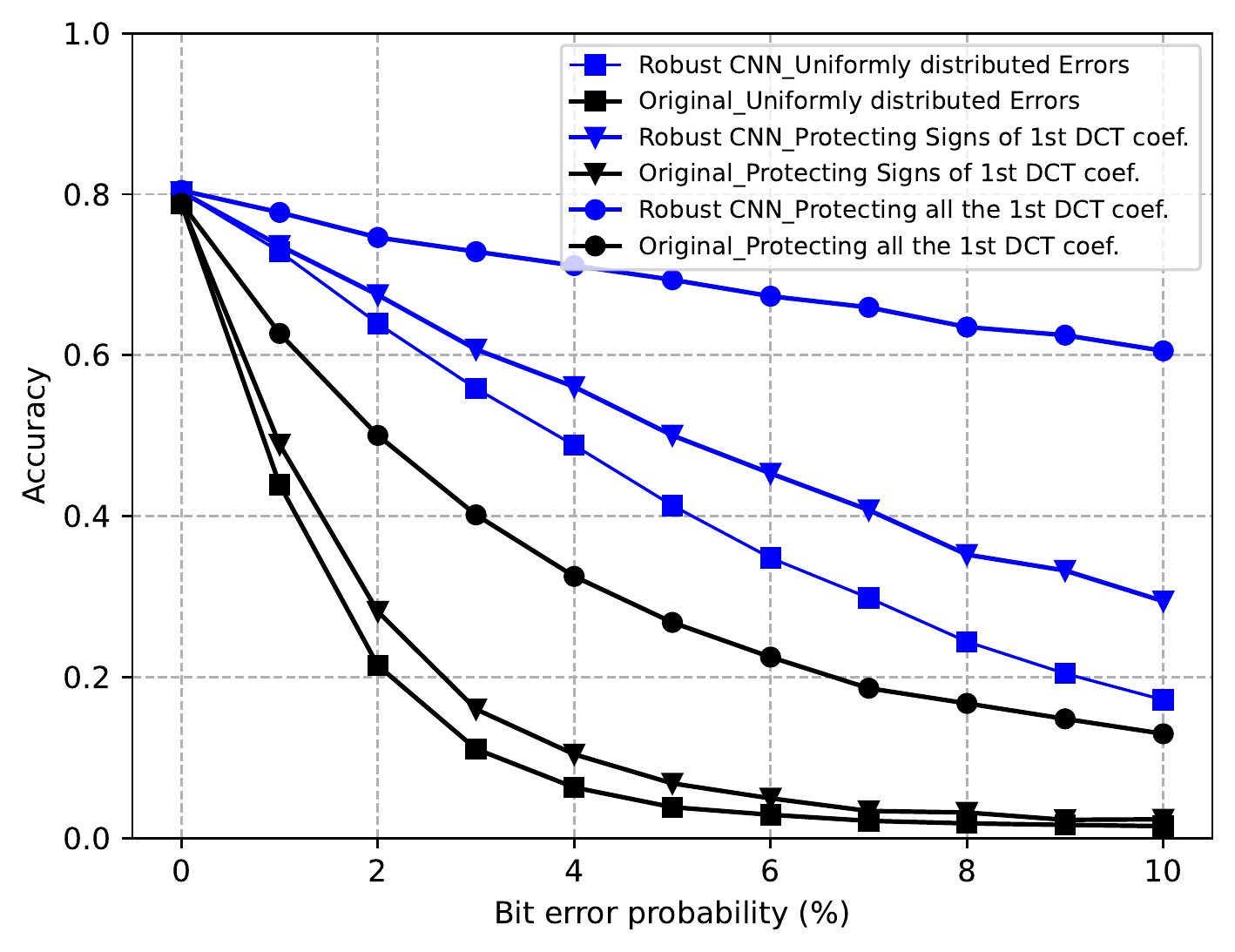}}
    \caption{Inference accuracy of MobileNet-V2 under different kinds of bit errors.}
    \label{fig:biterror}
\end{figure*}
\setcounter{table}{3}
\begin{table*}
\centering
  \caption{Average accuracy under 10\% bit error probability}
\label{tab:BitErrors}
  \begin{tabular}{ccccccc}
    \toprule
    \multicolumn{1}{c}{} & \multicolumn{3}{c}{\textbf{MobileNet-V2}} & \multicolumn{3}{c}{\textbf{ResNet-50}}\\
 & Original & Robust & Diff. & Original & Robust & Diff.\\ 
    \midrule
\textbf{Flower} &  &  &  &  \\ 
Uniformly distributed errors & 0.428 & \textbf{0.698} & +0.270 & 0.473 & \textbf{0.716} & +0.243 \\ 
Signs of the 1st DCT coef. protected & 0.478 & \textbf{0.748} & +0.270 & 0.502 & \textbf{0.783} & +0.281 \\
All the 1st DCT coef. protected & 0.669 & \textbf{0.861} & +0.192 & 0.714 & \textbf{0.877} & +0.163\\
\textbf{Caltech 256} &  &  &  &  \\ 
 Uniformly distributed errors & 0.159 & \textbf{0.445} & +0.286 & 0.149 & \textbf{0.442} & +0.293\\ 
Signs of the 1st DCT coef. protected & 0.186 & \textbf{0.520} & +0.334 & 0.175 & \textbf{0.516} & +0.341 \\
All the 1st DCT coef. protected & 0.342 & \textbf{0.696} & +0.354 & 0.337 & \textbf{0.720} & +0.383\\
  \bottomrule
\end{tabular}
\end{table*}
\subsection{Inference accuracy under bit errors}\label{subsec:biterrors}
During computation offloading, bit channel errors can occur. In this experiment, we study how bit errors and different effectiveness encoding strategies influence inference accuracy. First we tested Robust CNN models and the original models on images with uniformly distributed bit errors under different bit error probabilities. As shown in Fig.~\ref{fig:biterror}, the inference accuracy decreases as the bit error probability increases. The effectiveness decoding using Robust CNN models achieves higher inference accuracy (can also be seen in Table~\ref{tab:BitErrors}).

To improve inference accuracy under random bit errors, intuitively we can use FEC to protect the most important bits as one of the typical effectiveness encoding strategies. For binary representation, if the sign bit is flipped, it may cause significant change in values. In addition, as we know, the 1st DCT coefficients have the most important information about an image. The signs of the 1st DCT coefficients of an image are about less than 0.5\% of total data size of an image.
Therefore, we only need to add small amount of parity check bits, which increases transmission data size negligibly. As shown in Table~\ref{tab:BitErrors}, by protecting the signs of the 1st DCT coefficients, average inference accuracy of both the Robust models and their original models are improved by $2.6-7.5\%$ using the effectiveness encoding. For example, Original ResNet-50 improve the inference accuracy form 0.149 to 0.175 on Caltech 256. The effectiveness decoding Robust MobileNet-V2 and Robust ResNet-50 improve inference accuracy by 27\% and 28.1\% on flower dataset and 33.4\% and 34.1\% on Caltech 256, respectively. For example, Robust MobileNet-V2 achieves an average inference accuracy of 0.748 on flower dataset, while its original model only reaches 0.478.

To take a step further, since all the 1st DCT coefficients are about 4\% of the total image size, if protecting entire the 1st DCT coefficients using FEC, the increased transmission data size is still marginal; however, the inference accuracy is improved by $16.1-24.1\%$ on flower dataset and $33.8-40.7\%$ on Caltech 256 using this effectiveness encoding, as shown in Fig.~\ref{fig:biterror} and Table~\ref{tab:BitErrors}. For example, the average inference accuracy of Robust MobileNet-V2 on flower dataset is improved from 0.698 to 0.861 and that of its original model is improved from 0.428 to 0.669. In general, the Robust CNN models are shown to be robust against bit errors and consistently outperform their original models.

\subsection{Inference accuracy under latency and data rate constraints}\label{SubSec:delaytransmissionrateconstraints}
This section studies how the proposed Edge Intelligence framework performs under latency and data rate constraints, in comparison with the conventional framework. The data rate is assumed to be constant during task offloading. Note that the size of each image is not same even if using the same level of compression and same number of DCT coefficients. This is because the original image size depends on changes in image luminance and chrominance. We change the deadline and data rate in the range of $1-50$ ms and $1-50$ Mbps, respectively. As expected, the most critical conditions are those with low data rates and short deadlines. The simulations in this section are carried out using flower dataset in the following three scenarios.
\begin{figure}
    \centering
    \includegraphics[width=0.4\textwidth]{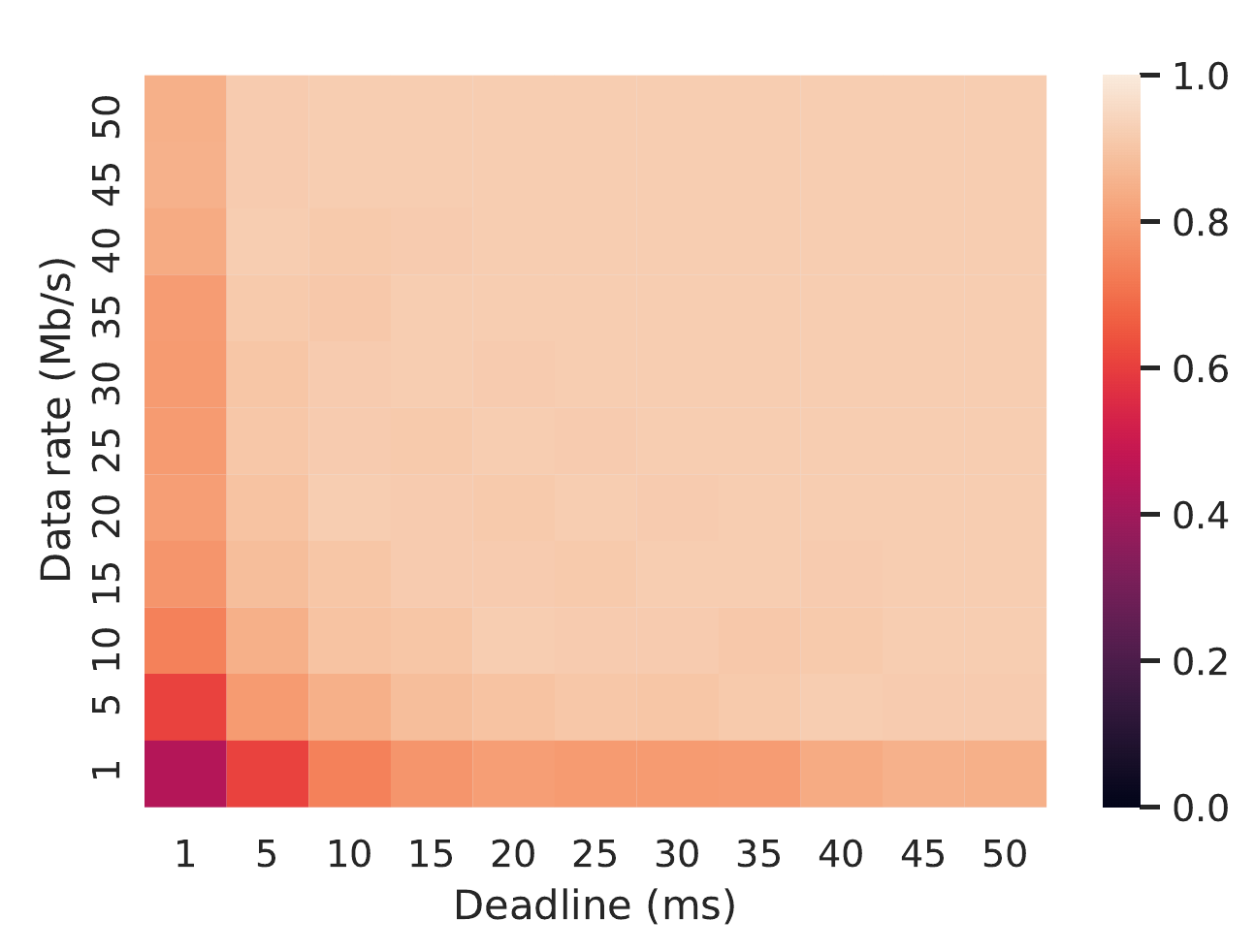}
    \caption{Inference accuracy of our Edge Intelligence framework with Robust ResNet-50 under latency and data rate constraints (No channel errors).}
    \label{fig:ourNoBitError}
\end{figure}
\begin{figure}
    \centering
    \includegraphics[width=0.4\textwidth]{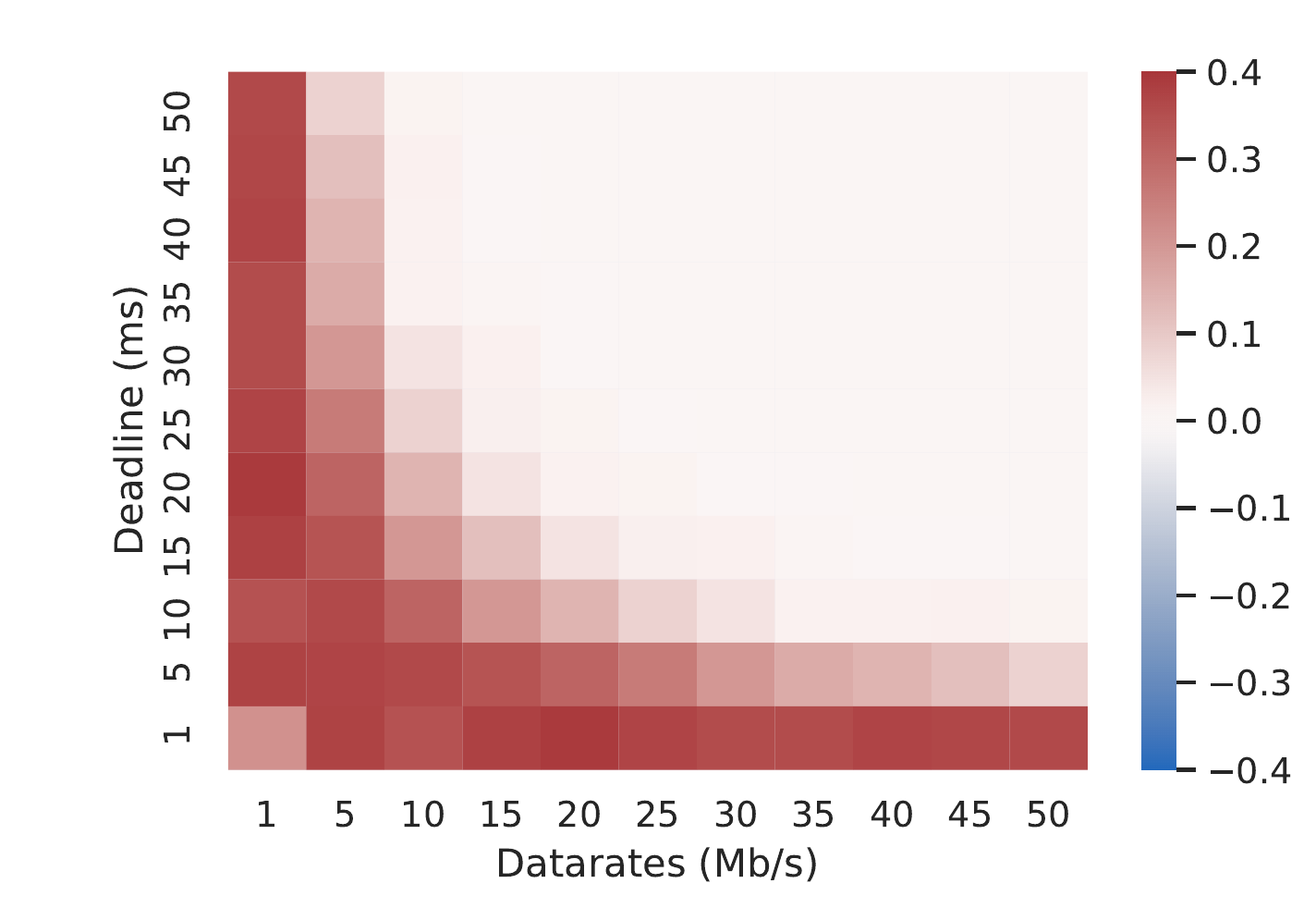}
    \caption{Difference in inference accuracy between our proposed Edge Intelligence framework (using Robust ResNet-50) and the conventional approach under latency and data rate constraints (No channel errors).}
    \label{fig:CompNoBitErrors}
\end{figure}
\subsubsection{No channel error}\label{nochannelerror}
Fig.~\ref{fig:ourNoBitError} shows the inference accuracy of the proposed Edge Intelligence framework using Robust ResNet-50 under latency and data rate constraints, when no channel error during offloading. It can be observed that the proposed framework can achieve high accuracy as long as the deadline is above 5 ms and data rate is above 5 Mbps. Fig.~\ref{fig:CompNoBitErrors} shows that the difference in the inference accuracy of the proposed framework and that of the conventional framework is small, if an image can be completely transmitted under the given deadline and data rate; however, under the critical conditions, namely, under stringent latency ($<15$ ms) and low data rate ($<25$ Mbps), our proposed framework performs significantly better, achieving a gain of inference accuracy up to 40\% using Robust ResNet-50.
\subsubsection{With channel errors}
\begin{figure}
    \centering
    \includegraphics[width=0.4\textwidth]{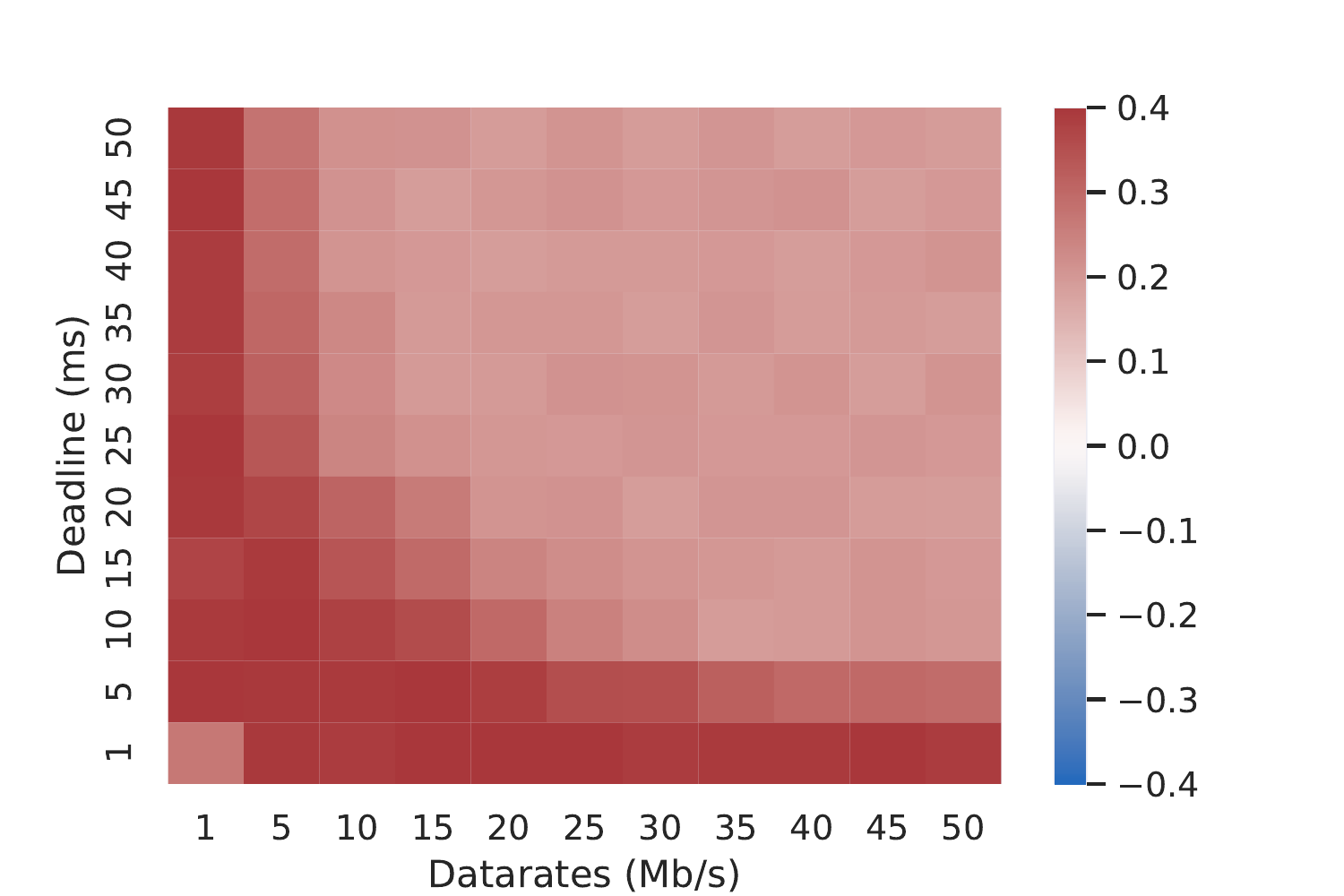}
    \caption{Difference in inference accuracy between our proposed Edge Intelligence framework (using ResNet-50) and the conventional approach with 5\% bit error probability and protecting entire the 1st DCT coefficients.}
    \label{fig:comparisionerror5}
\end{figure}

In this simulation, the bit error probability is set to 5\%. The latency and data rate constraints are the same as in Section \ref{nochannelerror}. According to the insights that obtained in Section~\ref{subsec:biterrors}, it is reasonable to protect entire the 1st DCT coefficients to minimize the impact on inference accuracy. As shown in Fig.~\ref{fig:comparisionerror5}, our proposed framework outperforms the conventional approach under channel errors and achieves a gain of inference accuracy up to 40\% using Robust ResNet-50. It is also clear that our proposed framework can achieve more gains in inference accuracy under more critical conditions (stringent latency and low data rate constraints).
\begin{figure}
    \centering
    \includegraphics[width=0.4\textwidth]{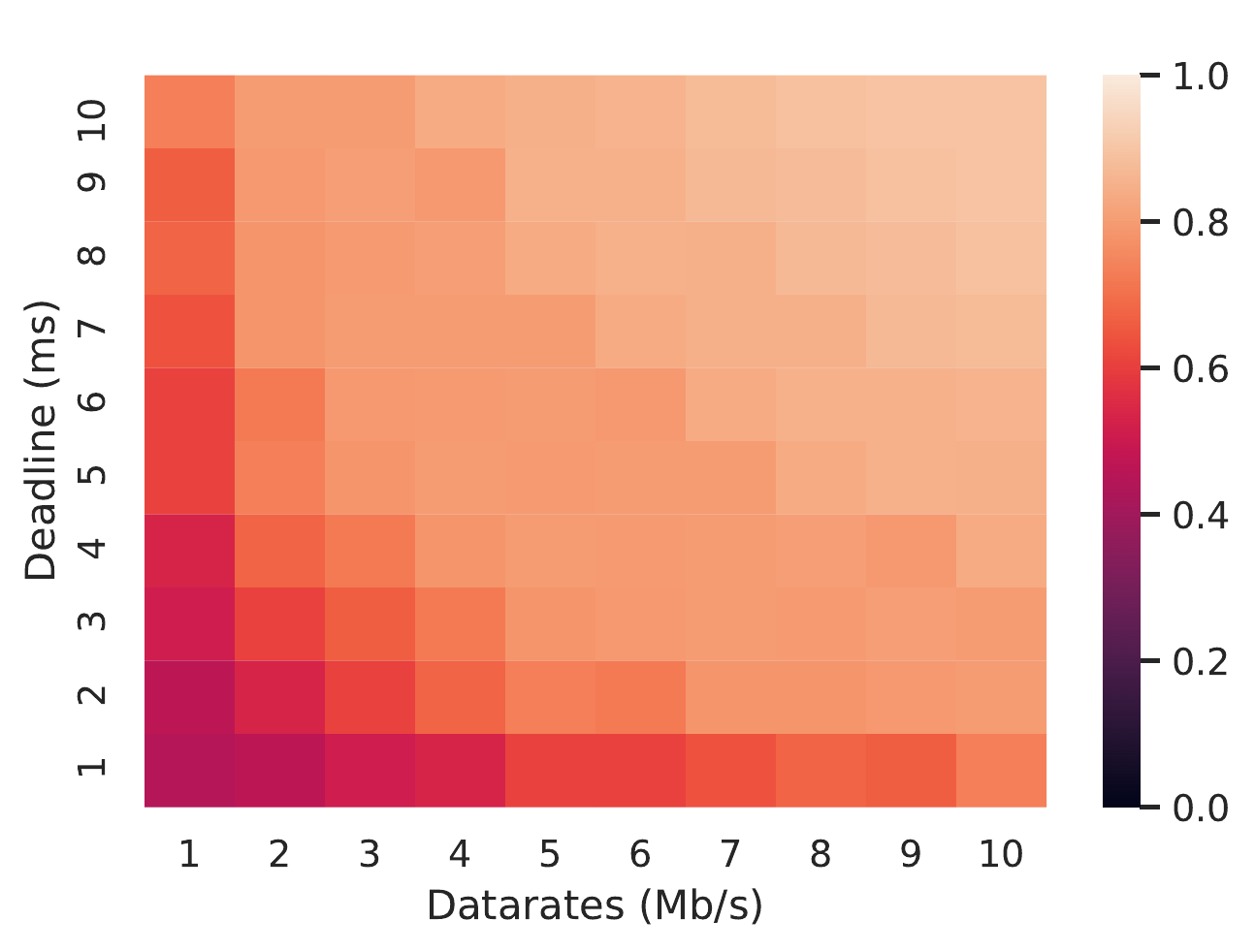}
    \caption{Inference accuracy of our Edge Intelligence framework (using ResNet-50) (deadline $< 10$ ms and data rate $<10$ Mbps).}
    \label{fig:stringentdeadlines}
\end{figure}
\subsubsection{Under extremely stringent deadline and low data rate}
When the latency constraint is below 10 ms and the available data rate is below 10 Mbps, the conventional approach fails to perform any inference. For these extremely critical scenarios, we test the performance our proposed framework plus additional protection of the 1st DCT coefficients in Fig.~\ref{fig:stringentdeadlines}, which shows that it can still achieve decent inference accuracy.
\section{Conclusion}\label{Section:Conclusion}
In this paper, we design an robust Edge Inference framework for time-critical IoT applications using semantic communication paradigm. We propose the channel-agnostic effectiveness encoding for offloading and realize an effectiveness decoding by implementing a novel image augmentation process for CNN training which generates Robust MobileNet-v2 and Robust ResNet-50 to provide high inference accuracy for distorted images. The experimental results show that the effectiveness decoding using Robust CNN models perform consistently better than their original models under different kinds of image distortions. Thanks to the design of semantic communication, our proposed Edge Intelligence framework significantly outperforms the conventional framework under latency and data rate constraints, in particular, under extremely stringent deadlines and low data rates. 
\bibliographystyle{IEEEtran}
\bibliography{IoT}

\end{document}